%% file: main.tex
\documentclass[runningheads]{llncs}
\usepackage{graphicx}
\usepackage{amsmath,amssymb}

\usepackage[width=122mm,left=12mm,paperwidth=146mm,height=193mm,top=12mm,paperheight=217mm]{geometry}
\usepackage{color}

\usepackage[table,xcdraw]{xcolor}
\usepackage{booktabs}
\usepackage{kotex}
\usepackage{multirow} 
\usepackage{subcaption}
\usepackage{pifont}
\newcommand{\cmark}{\ding{51}}
\newcommand{\xmark}{\ding{55}}

\begin{document}
\pagestyle{headings}
\mainmatter

\title{Vision Transformer-based Feature Extraction\\for Generalized Zero-Shot Learning}

\titlerunning{ViT-based Feature Extraction for GZSL}
\authorrunning{ViT-based Feature Extraction for GZSL}

\author{Jiseob Kim, Kyuhong Shim, Junhan Kim and Byonghyo Shim}
\institute{Department of Electrical and Computer Engineering, Seoul National University, Seoul, Korea \\
\small{\texttt{\{jskim, khshim, junhankim, bshim\}@islab.snu.ac.kr}}
}

\maketitle

\begin{abstract}
\input{sections/00_abstract}
\end{abstract}

\section{Introduction}\label{sec:intro}
\input{sections/01_introduction}

\section{Related Work}\label{sec:related}
\input{sections/02_related}

\section{ViT-GZSL}\label{sec:method}
\input{sections/03_method}

\section{Experiments}\label{sec:experiment}
\input{sections/04_experiment}

\section{Conclusion}\label{sec:conclusion}
\input{sections/05_conclusion}


\bibliographystyle{splncs}
\bibliography{egbib}


\appendix

\input{sections/a0_appendix} 

\end{document}

%% file: sections/00_abstract.tex
Generalized zero-shot learning (GZSL) is a technique to train a deep learning model to identify unseen classes using the image attribute. 
In this paper, we put forth a new GZSL approach exploiting Vision Transformer (ViT) to maximize the attribute-related information contained in the image feature. 
In ViT, the entire image region is processed without the degradation of the image resolution and the local image information is preserved in patch features. 
To fully enjoy these benefits of ViT, we exploit patch features as well as the CLS feature in extracting the attribute-related image feature.
In particular, we propose a novel attention-based module, called attribute attention module (AAM), to aggregate the attribute-related information in patch features.
In AAM, the correlation between each patch feature and the synthetic image attribute is used as the importance weight for each patch.
From extensive experiments on benchmark datasets, we demonstrate that the proposed technique outperforms the state-of-the-art GZSL approaches by a large margin.

%% file: sections/01_introduction.tex
Image classification is a long-standing yet important task with a wide range of applications such as autonomous driving, medical diagnosis, and industrial automation, to name just a few~\cite{autonomous_driving,industrial_automation,medical_diagnosis}.
In solving the task, deep learning-based supervised learning (SL) techniques have been popularly used~\cite{simonyan2014very,he2016deep}.
A well-known drawback of such SL is that it requires a large number of training samples for the accurate identification of all classes.
Unfortunately, in many practical scenarios, it is difficult to collect training samples for unpopular and rare classes (e.g., newly observed species or endangered species).
When there are \textit{unseen} classes where training data is unavailable, SL-based models are biased towards the \textit{seen} classes, deteriorating the identification of the unseen classes.

\begin{figure*}[t]
  \centering 
  \includegraphics[width=0.750\linewidth]{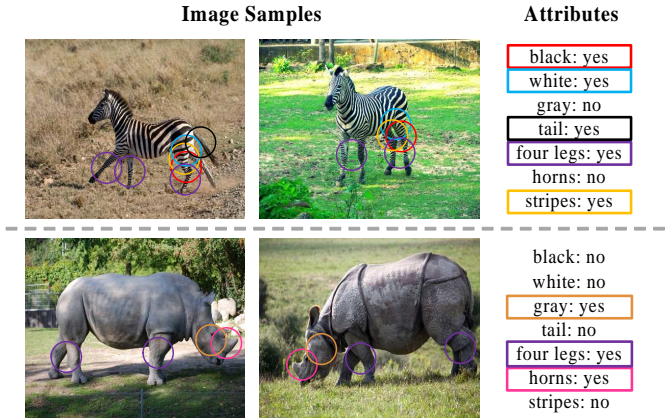}
  \caption{Examples of image attributes. The type of attributes are the same for all classes, but the values are different. Most of the attributes appear in a small region (e.g., horns, tail), while some attributes span a wide region (e.g., four legs, stripes).}
  \label{fig:attributes}
  \vspace{-10pt}
\end{figure*} 

\begin{figure*}[t]
  \centering 
  \includegraphics[width=0.85\linewidth]{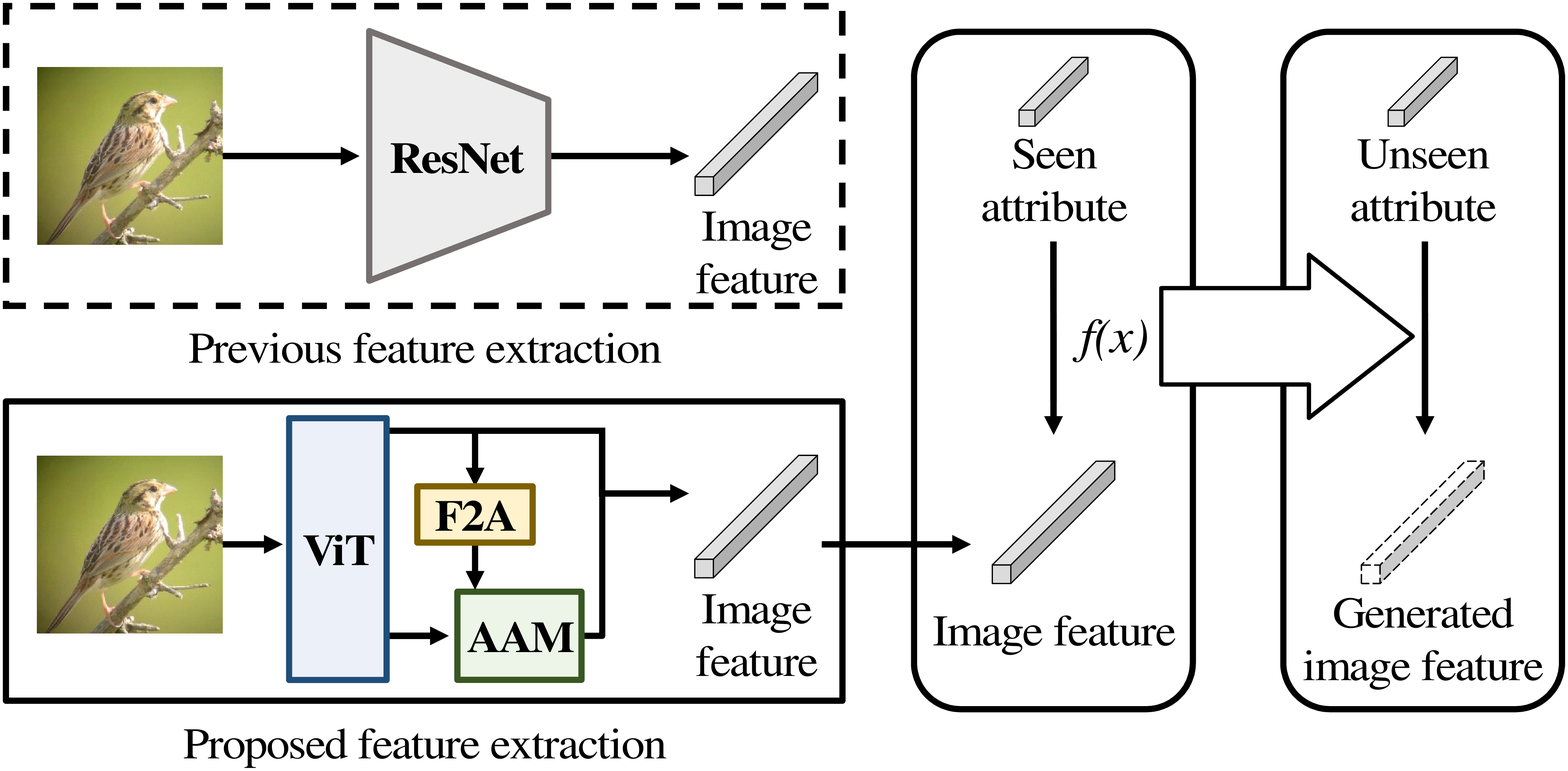}
  \caption{Illustration of our contribution. In short, we propose a ViT-based image feature extraction method to maximize the attribute-related information contained in the image feature.} 
  \label{fig:overview}
  \vspace{-5pt}
\end{figure*}

Recently, to overcome this issue, a technique to train a classifier using manually annotated attributes (e.g., color, size, and shape; see Fig.~\ref{fig:attributes}) has been proposed~\cite{zsl_proposal,gzsl_intro}.
Key idea of this technique, dubbed generalized zero-shot learning (GZSL), is to learn the relationship between the image feature and the attribute from seen classes and then use the trained model in the identification of unseen classes.
In~\cite{CVAE-GZSL,CLSWGAN}, the network synthesizing the image feature from the attribute has been employed to generate training data for unseen classes.
In extracting the image feature, a convolutional neural network (CNN) such as VGG~\cite{simonyan2014very} and ResNet~\cite{he2016deep} has been popularly used (see Fig.~\ref{fig:overview}). 

\begin{figure*}[t]
  \centering 
  \includegraphics[width=1.00\linewidth]{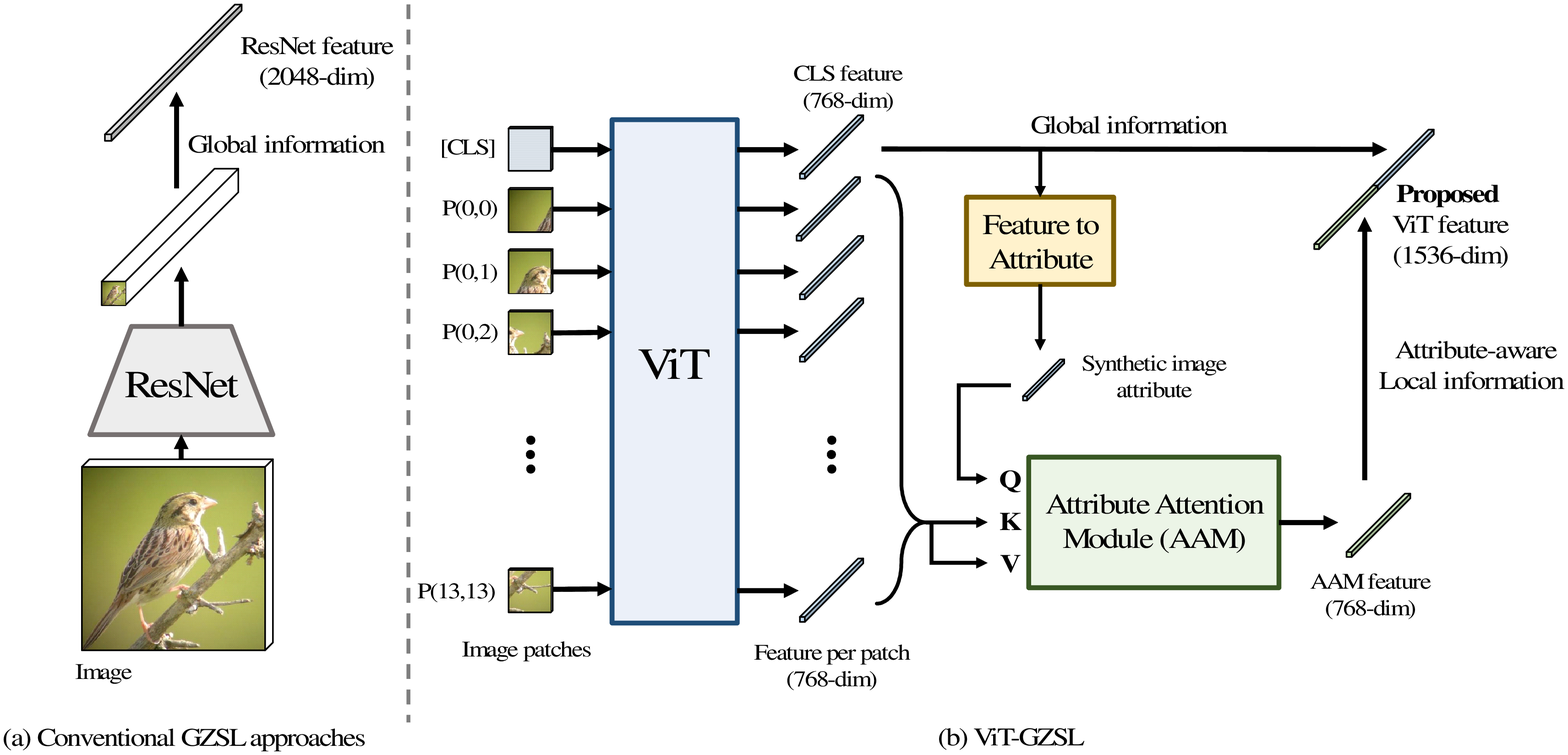}
  \caption{
  Illustration of image feature extraction methods in (a) conventional GZSL approaches and (b) the proposed ViT-GZSL.
  The image feature of ViT-GZSL is a concatenation of the CLS feature (Section~\ref{ssec:cls}) and the AAM feature (Section~\ref{ssec:aam}), which correspond to global information and attribute-aware local information, respectively.}
  \label{fig:concept}
  \vspace{-10pt}
\end{figure*}

A potential drawback of conventional CNN-based GZSL techniques is that image features might not contain all the attribute-related information. 
In fact, due to the locality of the convolution kernel, CNN requires a series of pooling and strided convolution operations to collect information of the whole image~\cite{DeepLearningBook}.
Since these operations lead to the spatial information loss~\cite{self_attention_GAN}, attribute-related features distributed in the whole area of an image (e.g., features corresponding to the `four legs' attribute; see Fig.~\ref{fig:attributes}) might not be extracted properly, affecting the learning quality considerably.

In this paper, we propose a new image feature extraction technique that enriches the attribute-related information in the extracted image feature.
Key ingredient of our approach is to employ Vision Transformer (ViT), a recently proposed transformer-based image feature extractor~\cite{dosovitskiy2020image}, to extract the attribute-related image feature.
Recently, ViT has received much attention for its excellent performance in various computer vision tasks such as image classification, object detection, and semantic segmentation, among others~\cite{DINO,DETR,fang2021you}.
The core component of ViT is the self-attention module that aggregates the information in all the image locations for the feature extraction~\cite{dosovitskiy2020image}. 
When compared to CNN, ViT can extract features spread over the whole image without degrading the image resolution so that the spatial loss, if any, caused by information skipping (e.g., max pooling, strided convolution, and global average pooling) can be prevented.
Indeed, it has been reported in~\cite{raghu2021vision} that ViT outperforms CNN in preserving spatial information.

In the proposed GZSL approach, henceforth referred to as ViT-GZSL, we extract the attribute-related image feature by exploiting all the output features obtained from ViT.
In ViT, a global image representation called \textit{CLS feature} and local image representations called \textit{patch features} are extracted from the input image (see Fig.~\ref{fig:concept}(b))~\cite{dosovitskiy2020image}.
Among these, the CLS feature is often used as an image feature for the image classification~\cite{fang2021you,DETR}.
A potential problem is that the CLS feature might not contain all the attribute-related information since the CLS feature is trained via the classification.
For this reason, the CLS feature may ignore the image details needed to distinguish different attributes (e.g., `hooked bill' and `curved bill' in the CUB dataset~\cite{WahCUB_200_2011}).
To supplement the attribute-related information not captured by the CLS feature, we exploit the patch features that preserve the local image information (e.g., `horns' attribute in Fig.~\ref{fig:attributes})~\cite{raghu2021vision}.
To aggregate the attribute-related information in the patch features, we propose a novel attention-based module called \textit{attribute attention module} (AAM).
Essence of AAM is to combine the patch features using the correlation between each patch feature and the image attribute as a weight.
Because the image attribute is unavailable for the inference stage, we jointly train a module called \textit{feature-to-attribute} (F2A) that generates a synthetic image attribute from the CLS feature.
From our experiments, we demonstrate that ViT-GZSL using both the AAM feature and the CLS feature outperforms ViT-GZSL only utilizing the CLS feature.

The main contributions of our work are summarized as follows:
\begin{itemize}
    \item We propose a new GZSL approach called ViT-GZSL.
    Key idea of ViT-GZSL is to use a ViT-based feature extractor to maximize the attribute-related information contained in the image feature.
    To the best of our knowledge, this work is the first one exploiting ViT in the feature extraction for GZSL. 
    
    \item We propose an attention-based architecture referred to as AAM that integrates the attribute-related information in the patch features.
    From the visualization of the attention weight, we demonstrate that AAM assigns a higher weight to the attribute-related image region.

    \item From extensive experiments on three benchmark datasets, we demonstrate that ViT-GZSL outperforms conventional GZSL approaches by a large margin.
    For example, for the CUB dataset, ViT-GZSL achieves 13.3\% improvement in the GZSL classification accuracy over the state-of-the-art technique~\cite{Composer}.
\end{itemize}

%% file: sections/02_related.tex
The main task in GZSL is to learn the relationship between image features and attributes and then use it in the identification of unseen classes.
Early GZSL works have focused on the training of a network measuring the compatibility score between image features and attributes~\cite{ALE,DeViSE}.
Once the network is trained properly, images can be classified into classes that have the most compatible attributes.  
In generative model-based approaches, synthetic image features of unseen classes are generated from their attributes~\cite{CVAE-GZSL,CLSWGAN}.
Popularly used generative models include the conditional variational autoencoder (CVAE)~\cite{VAE}, conditional generative adversarial network (CGAN)~\cite{CGAN}, and conditional Wasserstein GAN (CWGAN)~\cite{WGAN}.
By exploiting the generated image features of unseen classes as training data, a classification network can identify the unseen classes in a supervised manner.

Over the years, many efforts have been made to improve the performance of the generative model~\cite{CLSWGAN,LsrGAN,LisGAN,f-vaegan-d2,Zero-VAE-GAN,cycle-WGAN,DASCN,CADA-VAE,Disentangled-VAE,CE-GZSL}.
In~\cite{f-vaegan-d2,Zero-VAE-GAN,CADA-VAE,Disentangled-VAE}, an approach to combine multiple generative models (e.g., VAE and WGAN) has been proposed.
In~\cite{cycle-WGAN,DASCN}, an additional network estimating attributes from image features has been used to make sure that the generated image features satisfy the attributes of unseen classes.
In~\cite{CLSWGAN,LsrGAN,LisGAN}, an additional image classifier has been used in the training of the generative model to generate sufficiently distinct image features for different classes.
The key distinctive feature of the proposed technique over the previous studies is that we employ ViT as an image feature extractor to preserve the attribute-related information as much as we can.

%% file: sections/03_method.tex
In this section, we briefly go over the basics of ViT and discuss how to extract attribute-related image features using ViT.
We then delve into the GZSL classification using ViT-extracted features.

\subsection{Basics of ViT}\label{ssec:cls}

Inspired by the success of transformer-based models in natural language processing (NLP), various proposals applying Transformer to computer vision tasks have been made~\cite{chen2020generative,zhu2019text,chen2020image}.
Key idea of ViT is to split an input image into non-overlapping patches of the same size and then use the sequence of patches as an input to Transformer.
In ViT, the feature of each image patch is extracted using the self-attention mechanism in Transformer.
The self-attention mechanism is characterized by considering every patch that composes the whole image~\cite{vaswani2017attention}.
Since the entire image is used for the feature extraction, it is unnecessary to reduce the image resolution to increase the receptive field.
As a result, small image details can be well preserved in the features extracted by ViT.

We briefly walk through the notations used in ViT. 
Let $\mathbf{x}_{1}, \ldots, \mathbf{x}_{N}$ be $N$ non-overlapping patches of an input image.\footnote{If the input image size is $H \times W$ and the patch size is $P \times P$, then the total number of patches is $N={HW}/{P^2}$.}
As in transformer-based language models (e.g., BERT~\cite{BERT}), a special token $\mathbf{x}_{\text{CLS}}$ designed for the classification, called the CLS token, is appended at the front of the image patch sequence.
That is, $\mathbf{X}^{0} = [ \mathbf{x}_{\text{CLS}}; \mathbf{x}_{1}; \ldots; \mathbf{x}_{N} ]$ is used as an input sequence to Transformer (see Fig.~\ref{fig:concept}(b)).
Let $\mathbf{x}_{i}^{\ell}$ be the output of the $\ell$-th layer corresponding to $\mathbf{x}_{i}$ ($i \in \{ \text{CLS}, 1, \ldots, N \}$).
The outputs $\mathbf{x}_{\text{CLS}}^{\ell}$ and $\mathbf{x}_{i}^{\ell}$ are called the CLS feature and the patch feature of the $\ell$-th layer, respectively (see the supplementary material for the details of Transformer layer). 
Conventionally, the CLS feature of the last layer is used as a global image feature for downstream tasks~\cite{dosovitskiy2020image,DINO,touvron2021training}.
In this work, we exploit both CLS feature and patch features to better extract the attribute-related image feature.

\subsection{Attribute-related Feature Extraction for GZSL}\label{ssec:aam}  

\begin{figure*}[t]
  \centering 
  \includegraphics[width=1.00\linewidth]{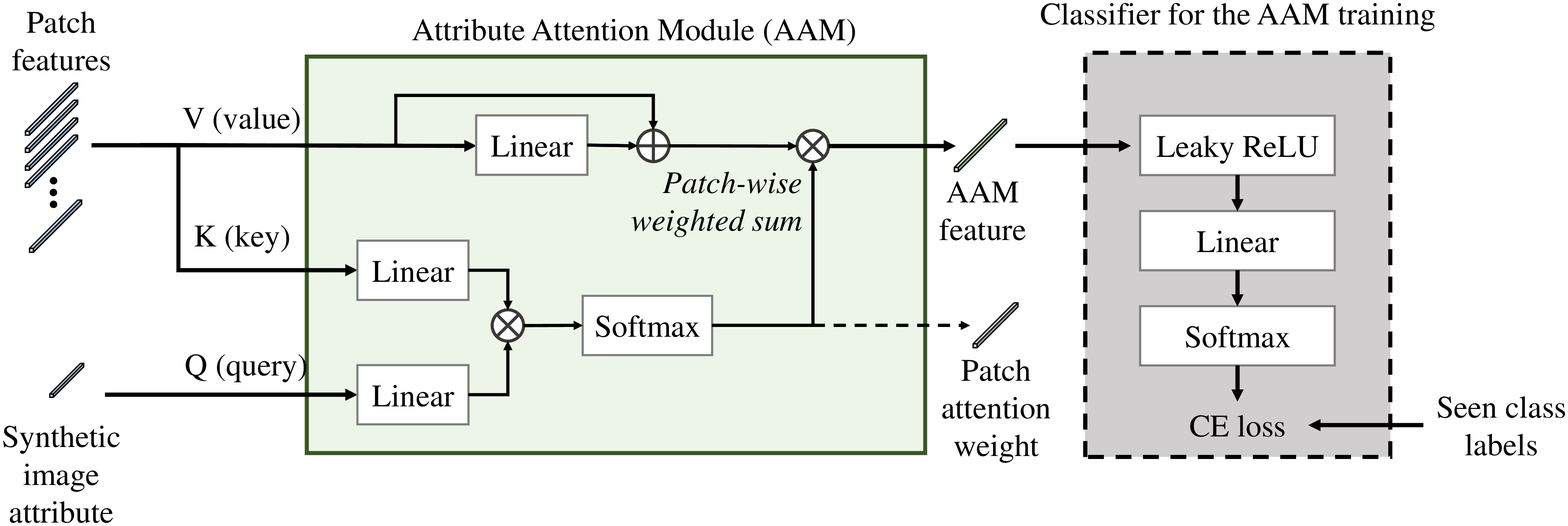}
  \caption{Illustration of the proposed attribute attention module (AAM). 
  AAM takes the patch features and the synthetic image attribute as inputs and extracts the attribute-related image feature by aggregating the patch features using the attention mechanism. CE loss indicates the cross-entropy loss. Multiple attention heads are omitted in the figure for simplicity.}
  \label{fig:aam_full}
  \vspace{-10pt}
\end{figure*}

\subsubsection{Main Idea}

Before discussing how to extract the attribute-related image information, we discuss why we choose ViT as an image feature extractor.
As illustrated in Fig.~\ref{fig:attributes}, there are two different types of image attributes: 1) attributes scattered in the entire image region (e.g., `four legs' attribute) and 2) attributes appearing only in a local image region (e.g., `tail' attribute). 
We use ViT in extracting these two types of attributes since ViT can process the entire image region without degrading the image resolution and at the same time preserve the local image information in the patch features~\cite{raghu2021vision}.
To make the most of the ViT, we need to use both the CLS feature and patch features in extracting the attribute-related image feature.

We note that not all patch features are related to attributes; as illustrated in Fig.~\ref{fig:attributes}, only a small number of image patches contain attribute-related important information.
To integrate essential patch features, we propose AAM, which combines patch features that are closely related to the image attribute.
Specifically, we take a weighted sum of patch features using the correlation between each patch feature and the image attribute as weight.
On the other hand, the real attributes of the given image are unknown during the inference stage.
To use AAM without real attributes, we propose feature-to-attribute (F2A) module to generate synthetic image attributes from the CLS feature. 
We illustrate the overall architecture in Fig.~\ref{fig:concept}.

\begin{figure*}[t]
  \centering 
  \includegraphics[width=0.65\linewidth]{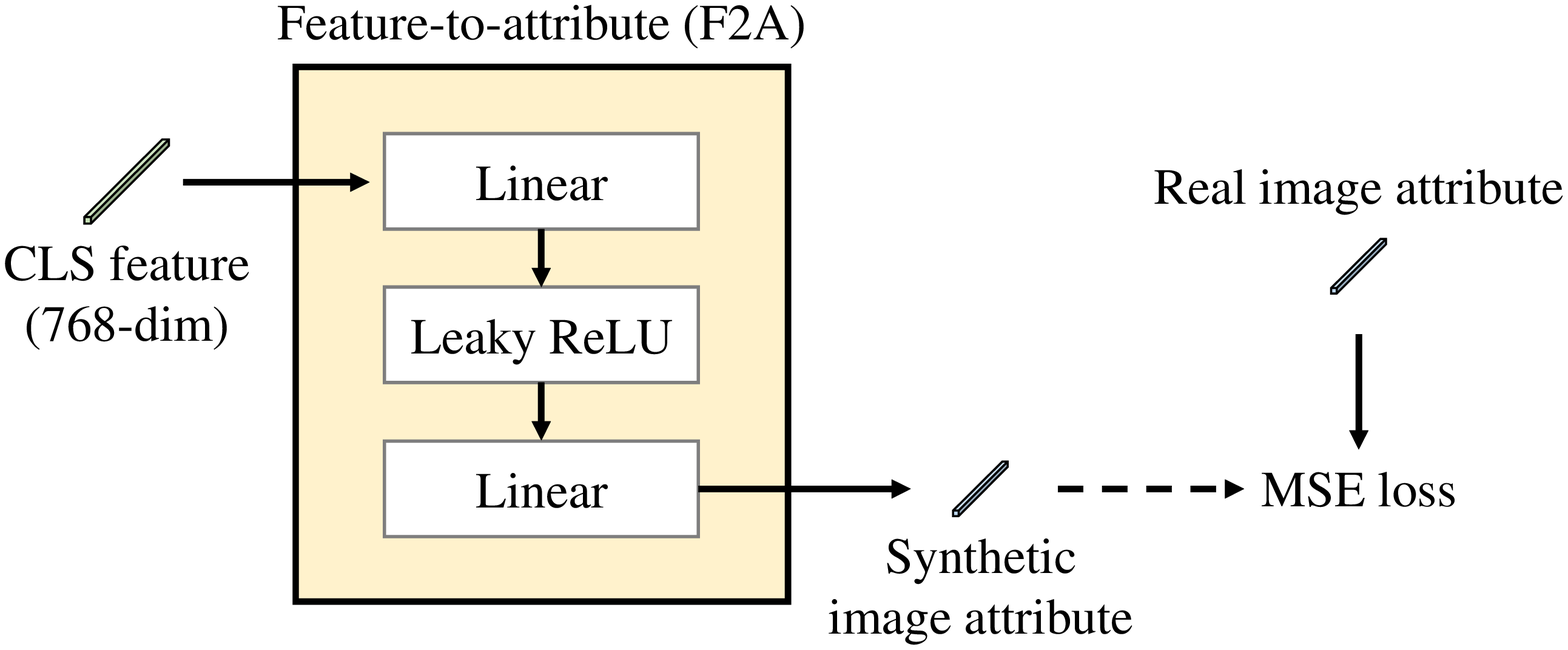}
  \caption{Illustration of the proposed feature-to-attribute (F2A) module. 
  This module takes CLS feature as input and generates the synthetic attribute. MSE is the mean square error. Real image attribute is only used during training.}
  \label{fig:f2a_full}
\end{figure*} 

\subsubsection{Details of F2A}\label{sssec:f2a} 
Before AAM, we generate synthetic attributes using a small 2-layer neural network, called feature-to-attribute (F2A) module.
The overall architecture of the F2A module is illustrated in Fig.~\ref{fig:f2a_full}.
To make the synthetic image attribute more realistic, we train the F2A module in a way to minimize the mean square error (MSE) between the true image attribute $\mathbf{{a}}$ and an estimated (synthetic) attribute $\mathbf{ \hat{a}}$.
We emphasize that the query of AAM is always a synthetic image attribute, for both training and inference stages.

\subsubsection{Details of AAM}

We discuss the detailed computation process of AAM.
Fig.~\ref{fig:aam_full} shows the architecture of AAM.
Similar to the cross-attention layer in Transformer~\cite{vaswani2017attention}, AAM takes query $Q$, key $K$, and value $V$ as inputs.
The output of AAM is a weighted sum of each row in $V$, where the (attention) weight is computed as the correlation between $Q$ and $K$.
In the proposed AAM, we use the image attribute for $Q$ and patch features $\mathbf{X}_{\text{patch}}^{\ell} = [\mathbf{x}_1^{\ell} ; \, \mathbf{x}_2^{\ell} ; \ldots; \, \mathbf{x}_N^{\ell} ]$ for $K$ and $V$:
\begin{align}
    Q &= \text{Linear}_Q(\mathbf{ \hat{a}}) \quad &\in \mathbb{R}^{1 \times d}, \label{eq:query} \\
    K &= \text{Linear}_K(\mathbf{X}_{\text{patch}}^{\ell}) \quad &\in \mathbb{R}^{N \times d},\\
    V &= \mathbf{X}_{\text{patch}}^{\ell} + \text{Linear}_V(\mathbf{X}_{\text{patch}}^{\ell}) \quad &\in \mathbb{R}^{N \times d}.\label{eq:value}
\end{align}
We add the residual connection from $\mathbf{X}_{\text{patch}}^{\ell}$ to $V$ since we empirically observed that AAM is highly overfitting on the seen classes in the absence of the residual connection.
To execute weighted sum of each patches, we calculate the patch attention weight $A_{p}$ as a scaled dot-product of $Q$ and $K$ followed by the softmax function:
\begin{equation}
    A_{p} 
    = \text{Softmax} \left ( \frac{{Q} {K}^T}{\sqrt{d}} \right ) .
\end{equation} 
Next, the patch features are combined using the attention weight $A_{p}$.
The AAM feature, an output of AAM, is given by
\begin{equation}
    \mathbf{x}_{\text{AAM}}^{\ell} = A_{p}V .
\end{equation}
Note that we exploit the multi-head version of attention computation for experiments, following the common practice~\cite{vaswani2017attention}.

To make AAM pays more attention to attribute-related patches, we exploit an auxiliary classifier that identifies the image class from the AAM feature (see the dashed box in Fig.~\ref{fig:aam_full}).
Specifically, we train AAM with the cross-entropy (CE) loss such that the classifier can predict the correct class label of an input image from the AAM feature.
We jointly train F2A module with AAM by minimizing the total loss, which is the sum of CE loss (for AAM) and MSE loss (for F2A).
Note that AAM and F2A are only trained with seen class images and attributes.
Also, the backbone ViT is initialized with pre-trained parameters and \textit{not fine-tuned} for the entire experiment.
After the AAM training, we generate the ViT feature $\mathbf{x}_{\text{ViT}}$ by stacking the AAM feature and the CLS feature and then use it as our image feature:
\begin{equation}
    \mathbf{x}_{\text{ViT}} 
    = [\mathbf{x}_{\text{CLS}}^{\ell_{1}}, \mathbf{x}_{\text{AAM}}^{\ell_{2}}] .
\label{eq:vit_concat}
\end{equation} 
Note that the layer index $\ell_{1}$ and $\ell_{2}$ may not be the same for the CLS feature and AAM feature.

\subsection{GZSL Classification Using ViT Features} 
 
So far, we have discussed how to extract attribute-related image features using ViT. 
Now, we discuss how to perform the GZSL classification using the ViT feature.
In a nutshell, we synthesize ViT feature samples for unseen classes from their attributes.
Once the synthetic samples for unseen classes are generated, the classifier identifying the image class from the ViT feature is trained in a supervised manner.\footnote{ViT feature samples for seen classes can be obtained from training image samples by exploiting pre-trained ViT and the proposed F2A+AAM modules.}
For the synthetic ViT feature generation, we use conditional variational autoencoder (CVAE)~\cite{CVAE-GZSL} as our generative model.

\subsubsection{ViT Feature Generation Using CVAE}

In synthesizing ViT features from image attributes, CVAE employs two networks, an encoder and a decoder, which are trained to maximize the conditional likelihood $p(\mathbf{x}_{\text{ViT}} | \mathbf{a})$~\cite{VAE}. 
In the training phase, the encoder maps the ViT feature $\mathbf{x}_{\text{ViT}}$ and the image attribute $\mathbf{a}$ into a random Gaussian latent vector $\mathbf{z} \sim \mathcal{N}(\mathbf{\mu}_{\mathbf{z}}, \mathbf{\Sigma}_{\mathbf{z}})$.
Then, the decoder reconstructs the original ViT feature $\mathbf{x}_{\text{ViT}}$ using the latent vector $\mathbf{z}$ and the image attribute $\mathbf{a}$. 
Let $\widehat{\mathbf{x}}_{\text{ViT}}$ be the reconstructed ViT feature, then the encoder and decoder are jointly trained to minimize the evidence lower bound (ELBO) of the conditional log likelihood $\log p(\mathbf{x}_{\text{ViT}} | \mathbf{a})$~\cite{VAE}.
Please see the supplementary material for the details.
After the CVAE training, we use the decoder in synthesizing ViT features of unseen classes.
Specifically, for each unseen class $u$, we generate synthetic ViT features $\widetilde{\mathbf{x}}_{\text{ViT}}$ by sampling a latent vector $\mathbf{z} \sim \mathcal{N}(\mathbf{0}, \mathbf{I})$ and then exploiting $\mathbf{z}$ and the image attribute $\mathbf{a}_{u}$ as inputs to the decoder.
By re-sampling the latent vector, a sufficient number of synthetic ViT features can be generated.

While we used CVAE, perhaps the most basic generative model, as a feature generation network, the proposed ViT-based feature can be integrated with any other advanced generative model such as f-CLSWGAN and TF-VAEGAN.
The reason why we used the simple CVAE model is to emphasize the importance of a good image feature extractor in GZSL.

\subsubsection{ViT Feature-based Classification}
After generating synthetic ViT feature samples for unseen classes, we train the ViT feature classifier using a supervised learning model (e.g., softmax classifier, support vector machine, and $K$-nearest neighbor).
In this work, to compare the ViT feature directly with the CNN-based image feature, we use the simple softmax classifier as a classification model.
Please see the supplementary material for the details of the classifier.

%% file: sections/04_experiment.tex
\subsection{Setup}\label{ssec:setup}

\subsubsection{Dataset and Metric} 
We evaluate the performance of our model using three benchmark datasets\footnote{We omit the commonly used AWA1 dataset because its raw images are not available anymore due to copyright restrictions.}: Animals with Attributes 2 (AWA2)~\cite{XianAWA2}, CUB-200-2011 (CUB)~\cite{WahCUB_200_2011}, and SUN attribute (SUN)~\cite{patterson2012sun}.
The AWA2 dataset contains 50 classes (seen 40 + unseen 10) of animal images annotated with 85 attributes.
The CUB dataset contains 200 species (150 + 50) of bird images annotated with 312 attributes.
The SUN dataset contains 717 classes (645 + 72) of scene images annotated with 102 attributes.

We adopt the conventional dataset split and the standard GZSL evaluation metric presented in~\cite{XianAWA2}.
Specifically, we measure the average top-1 classification accuracies $acc_{s}$ and $acc_{u}$ on seen and unseen classes, respectively, and then use their harmonic mean $acc_{h} = (2 \cdot acc_{s} \cdot acc_{u})/(acc_{s} + acc_{u})$ as a metric to evaluate the performance of GZSL methods.

\begin{figure}[t]
    \centering 
    \includegraphics[width=0.99\linewidth]{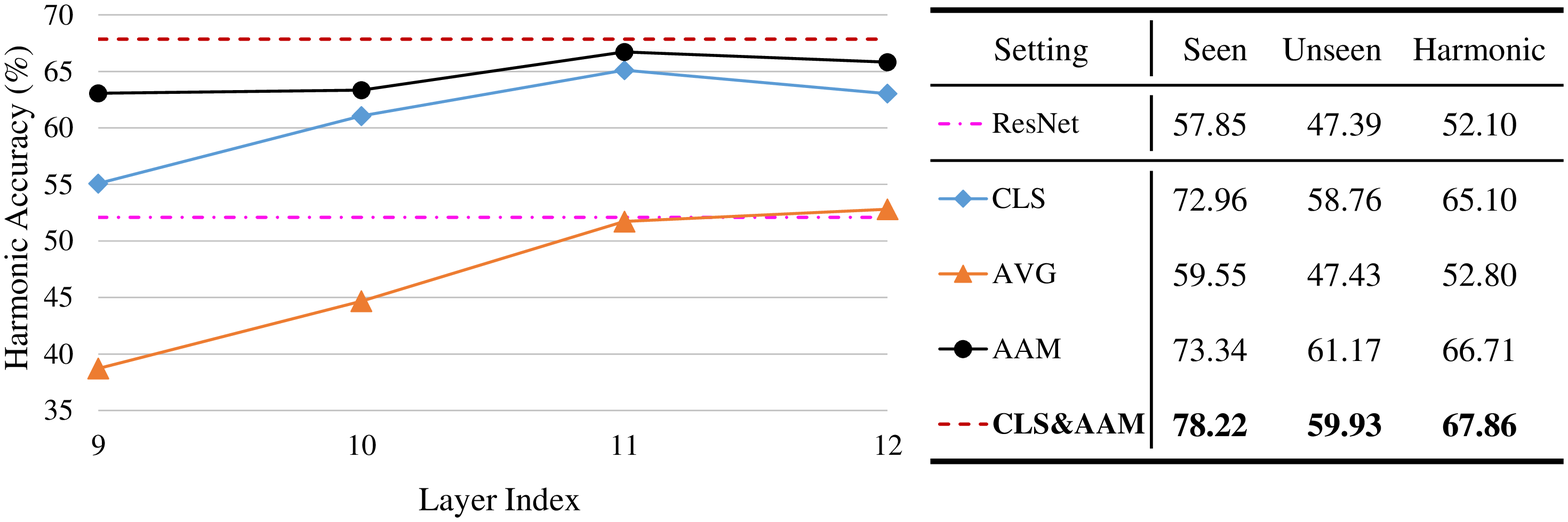}
    \caption{GZSL performance evaluated on the CUB dataset.
    Left figure shows results with respect to different Transformer layers used in the feature extraction. 
    Right table shows a detailed comparison of different image features. 
    }
    \label{fig:layer_accuracy_table}
\end{figure} 

\subsubsection{Implementation Details} 
We choose the DeiT-Base~\cite{touvron2021training} model, a variant of the original ViT model, as our feature extractor. Please note that the structure of the DeiT model is identical to the original ViT model, and it is widely used as a backbone for many previous works~\cite{chen2022transformers,el2022pain}.
As in ResNet-101~\cite{he2016deep}, DeiT-Base is trained on ImageNet-1K~\cite{Imagenet} for the image classification task with the image resolution of $224\times224$.
We employ the open-source pre-trained parameters\footnote{\texttt{https://huggingface.co/facebook/deit-base-distilled-patch16-224}} of the model.
Note that we use the fixed pre-trained backbone parameters in the training process.
In other words, only AAM and F2A modules are trained.
We use multi-head attention for AAM where the number of attention heads is set to 6 (attention head dimension is $128=768/6$).
In computing the query, key, and value for AAM (see~\eqref{eq:query}-\eqref{eq:value}), we use heavy dropout regularization after the linear projection to prevent the overfitting on the seen classes. 
As in conventional GZSL approaches~\cite{CVAE-GZSL,CADA-VAE,f-vaegan-d2}, we implement the encoder and decoder in CVAE using multi-layer perceptron (MLP) with one hidden layer.
We set the number of hidden units to 4096 and use leaky ReLU as a nonlinear activation function, following~\cite{f-vaegan-d2}.

\subsection{ViT Layer Selection and Feature Comparison}\label{ssec:layer_selection}

We first investigate which Transformer layer of ViT provides the most informative features.
Although we usually utilize the CLS feature of the last layer for many downstream tasks, we find that other layers may show better GZSL performance.
Fig.~\ref{fig:layer_accuracy_table} shows the effect on the accuracy by changing the feature extraction layer.
The higher accuracy implies that the extracted feature is more beneficial in terms of GZSL performance.
We omit the results for the first 8 layers since the corresponding CLS and patch features fail to achieve comparable performance.
We compare three different features; CLS ($\mathbf{x}_{\text{CLS}}$), AVG ($\mathbf{x}_{\text{AVG}}$), and AAM ($\mathbf{x}_{\text{AAM}}$), where $\mathbf{x}_{\text{AVG}}$ is a simple average of all patch features.
\begin{equation}
    \mathbf{x}_{\text{AVG}}^l = \frac{1}{N}\sum_{i=1}^{N} \mathbf{x}_i^l .
\vspace{3pt}
\end{equation}

First, we observe that the 11-th layer (total 12 layers) shows the best performance for the CLS feature.
The CLS feature of the last layer is specially trained for the classification task during pre-training; therefore, it may not obtain the attribute-related information unused to classify the images in the pre-training dataset (i.e., ImageNet-1K).
We can prevent this potential information loss by selecting lower layers other than the last layer.
Our result shows that the image feature required for GZSL is different from the image feature used in the classification.

\input{sections/04_experiment_result}

Second, we observe that the AVG feature shows much worse performance than the AAM feature.
Even though the AAM feature and AVG feature take the same patch features as input, the performance gap is significant, which verifies the attribute-related information contained in the patch features should be combined carefully.
Our AAM successfully integrates the patch features by assigning larger weights to attribute-related patches.

Table in Fig.~\ref{fig:layer_accuracy_table} reports the accuracy of features obtained from the layer with the best performance.
Compared to ResNet-based feature, the combination of CLS feature and AAM feature (proposed ViT feature) achieve more than 30\% improvement in harmonic accuracy. 
By merging both features, we can further improve the accuracy compared to using only CLS feature or AAM feature, which implies that the two features contain complementary attribute-related information.

\subsection{Comparison with State-of-the-art}\label{ssec:result}

\subsubsection{GZSL Performance} 
In Table~\ref{tab:result}, we summarize the performance of the proposed ViT-GZSL technique and conventional GZSL approaches on different datasets.
From the results, we observe that for all datasets, ViT-GZSL outperforms conventional CNN-based approaches by a large margin.
Specifically, for the CUB dataset, ViT-GZSL achieves more than 13\% improvement in the harmonic accuracy over the state-of-the-art approach.
Also, for the AWA2 and SUN datasets, the harmonic accuracy of ViT-GZSL is at least 5\% and 13\% higher than those obtained by the state-of-the-arts, respectively.

We would like to compare our work with two groups of works that aim to improve the quality of the image feature: 1) approaches removing the attribute-irrelevant information from the image feature (e.g., DLFZRL~\cite{DLFZRL}, Disentangled-VAE~\cite{Disentangled-VAE}, RFF-GZSL~\cite{RFF-GZSL}, and SE-GZSL~\cite{kim2021semantic}) and 2) approaches extracting the image feature of each individual attribute by exploiting additional attribute information (e.g., DAZLE~\cite{DAZLE} and Composer~\cite{Composer}).
The harmonic accuracy of ViT-GZSL on the CUB dataset is 20\% higher than those obtained by the first group of works.
Also, ViT-GZSL achieves more than 46\% improvement on the SUN dataset over the second group of works.

The performance gap between ViT-GZSL and the conventional approaches is because 1) the attribute-related information loss caused by the degradation of the image resolution in CNN is prevented by exploiting ViT in our approach and 2) the attribute-related information, not captured by the global image feature (CLS feature in our approach and ResNet feature in conventional approaches), can be extracted by exploiting the additional patch features in our approach.

\setlength{\tabcolsep}{8pt}
\begin{table*}[t]
    \centering
    \caption{Performance comparison on the effect of MSE loss for F2A.}
    \vspace{0.3cm}
    \begin{tabular}{c | ccc | ccc}
        \toprule
        \multirow{2}{*}{MSE Loss} &  \multicolumn{3}{c|}{AAM Feature} &  \multicolumn{3}{c}{CLS \& AAM Feature} \\ 
                 & $acc_{s}$ & $acc_{u}$ & $acc_{h}$ & $acc_{s}$ & $acc_{u}$ & $acc_{h}$ \\ 
        \midrule
        \xmark     & 54.95    & 48.98    & 57.79    & 75.18 & 52.43 & 61.78    \\ 
        \cmark     & 73.34    & 61.17    & \textbf{66.71}    & 78.22 & 59.93 & \textbf{67.86}    \\ 
        \bottomrule
    \end{tabular}
    \label{tab:f2a_anaysis}
\vspace{-10pt}
\end{table*}
\setlength{\tabcolsep}{2pt}

\subsubsection{Importance of Feature Extractor} 

We would like to mention that ViT-GZSL is identical to CVAE-GZSL$^\dagger$ except for the fact that ViT-GZSL uses a ViT-based feature extractor while CVAE-GZSL$^\dagger$ exploits a CNN-based feature extractor.
From the results in Table~\ref{tab:result}, we show that the only difference in the feature extractor leads us to about 11\%, 30\%, and 22\% improvement in the harmonic accuracy on the AWA2, CUB, and SUN datasets, respectively.

In conventional generative model-based approaches, various networks (e.g., attribute classifier~\cite{cycle-WGAN}, discriminator~\cite{f-vaegan-d2}, and feedback module~\cite{TF-VAEGAN}) are used to improve the performance of generative models.
In our approach, we do not include such networks in the generative model training to clearly show the performance improvement coming from the ViT-based feature extractor.
We expect that the performance of ViT-GZSL can be improved further by exploiting an advanced generative model, e.g., TF-VAEGAN~\cite{TF-VAEGAN}.
Since the focus of this paper is on the importance of a good image feature extractor, we leave the exploration of advanced generative models as future work.

\subsection{Analysis}

\subsubsection{Training Loss for F2A}\label{ssec:f2a}

To understand the synthetic attribute generated from F2A, we conduct an ablation study on the MSE loss during the AAM training.
Remind that MSE loss is used to enforce the output of F2A (synthetic attribute) to be more likely to the real image attributes.
Table~\ref{tab:f2a_anaysis} shows the effect of MSE loss on CUB dataset.
Without MSE loss, the GZSL performance is significantly degraded and the output of F2A becomes very different from the real ones.
This implies that it is very important to provide proper class-related attributes for the query of AAM in extracting semantically meaningful information from ViT patch features.

\subsubsection{Patch Attention Map from AAM}\label{ssec:aam_attention}

\begin{figure*}[t]
  \centering 
  \includegraphics[width=0.92\linewidth]{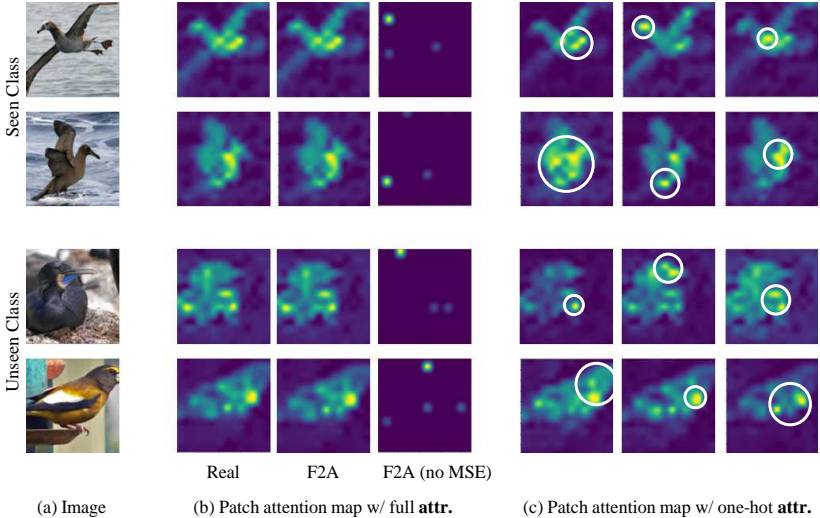}
  \caption{Visualization of patch attention weights from AAM. 
  The first two rows are from the seen classes and the last two rows are from the unseen classes.
  The second, third, and fourth columns (b) represent the patch attention map from the real and synthetic image attributes from F2A and synthetic image attributes from F2A without MSE loss, respectively.
  The last three columns (c) show exemplar attention maps corresponding to one-hot image attributes.
  The white circles on the last three columns indicate large-valued attention locations.
  A brighter area indicates a higher probability. For better visualization, the attention map is min-max normalized. Best viewed in color.}
  \label{fig:aam_attention}
  \vspace{-5pt}
\end{figure*}

To verify that AAM actually focuses on the attribute-related image location, we visualize the patch attention map\footnote{Attention map is averaged through multiple attention heads for the visualization.} $A_p$ of AAM.
Remind that each element of the attention map indicates how much AAM pays attention to a certain location of the image.

First, we visualize the attention map generated from the real image attribute ($2^{\text{nd}}$ column in Fig.~\ref{fig:aam_attention}).
We can observe that AAM successfully attends to the object region for both seen and unseen images.
This behavior supports our two key statements: 1) AAM learns to focus on attribute-related regions, and 2) patch features of ViT contain attribute-related information.
This explains the failure of the AVG feature in that only a limited number of patches contain the high probability (highly correlated to the attribute) area.

Second, we show the attention map using the synthetic image attribute generated from F2A module ($3^{\text{rd}}$ column in Fig.~\ref{fig:aam_attention}).
The resulting attention map is almost identical to the ones that used the real image attribute, which implies that F2A module successfully learned to synthesize the attribute from CLS feature although F2A is only trained with seen class images.
We also show the attention map in the absences of MSE loss ($4^{\text{th}}$ column in Fig.~\ref{fig:aam_attention}).
Without MSE loss, AAM fails to find attributes in the image.

\setlength{\tabcolsep}{6pt}

\begin{table}[t]
    \centering
    \caption{Example of the generated one-hot image attributes.
    One-hot attributes only contain a single non-zero value (10.0) for the selected attribute.}
    \vspace{0.2cm}
    \resizebox{0.8\linewidth}{!}{
    \begin{tabular}{c|ccccccc}
        \toprule
        \multirow{2}{*}{Attribute}  & bill:  & wing: & tail:   & ... & back: & eye:  & forehead: \\
                                    & dagger & black & striped & ... & red   & olive & black \\
        \midrule
        Real                        & 0 & 0.31 & 0.76 & ... & 0.04 & 0 & 0.88 \\  
        Case 1                      & 0 & 0 & 10.0 & ... & 0 & 0 & 0 \\
        Case 2                      & 0 & 0 & 0 & ... & 0 & 0 & 10.0 \\
        \bottomrule
    \end{tabular}}
    \label{tab:example_synthetic}
\vspace{-5pt}
\end{table}
\setlength{\tabcolsep}{8pt}

Finally, we examine whether AAM actually learns to localize each attribute.
We expect that AAM should generate a different attention map for a different image attribute even if the input image is unchanged. 
We choose several high-valued attributes within the image attribute and create the one-hot attributes containing a single attribute (see Table~\ref{tab:example_synthetic} for some examples).
Since one-hot attributes only include a single attribute of the image, we expect that AAM pays more attention to the corresponding attribute when these one-hot attributes are given as queries.
The last three columns in Fig.~\ref{fig:aam_attention} present the attention map generated from selected one-hot attributes.
As expected, AAM attends to different parts of the image for different attributes.
Recalling that the goal of GZSL is to learn the relationship between image attributes and features, the ability to learn each attribute and localize the corresponding part in the image supports the superior performance of our approach.

%% file: sections/04_experiment_result.tex
\setlength{\tabcolsep}{4pt}
\begin{table*}[t]
    \centering
    \caption{GZSL performance of the proposed ViT-GZSL and conventional approaches. 
    $acc_{s}$, $acc_{u}$, and $acc_{h}$ are the seen, unseen, and harmonic accuracy (\%), respectively.
    The best results are in bold and the second best results are underlined.
    For a fair comparison, we also report the performance of CVAE-GZSL exploiting the softmax classifier (see CVAE-GZSL$^\dagger$).}
    \vspace{0.5cm}
    {
    \begin{tabular}{c|ccc|ccc|ccc}
    \toprule
    \multirow{2}{*}{Model} & \multicolumn{3}{c|}{AWA2} & \multicolumn{3}{c|}{CUB} & \multicolumn{3}{c}{SUN} \\
    \cline{2-10}
    & $acc_{s}$ & $acc_{u}$ & $acc_{h}$ & $acc_{s}$ & $acc_{u}$ & $acc_{h}$ & $acc_{s}$ & $acc_{u}$ & $acc_{h}$ \\ 
    \midrule
    CVAE-GZSL~\cite{CVAE-GZSL} & - & - & 51.2 & - & - & 34.5 & - & - & 26.7  \\ \hline
    CVAE-GZSL$^\dagger$ & 70.9 & 60.1 & 65.1 & 57.9 & 47.4 & 52.1 & 36.1 & 45.0 & 40.1 \\ \hline
    f-CLSWGAN~\cite{CLSWGAN} & - & - & - & 57.7 & 43.7 & 49.7 & 36.6 & 42.6 & 39.4 \\ \hline
    cycle-CLSWGAN~\cite{cycle-WGAN} & - & - & - & 61.0 & 45.7 & 52.3 & 33.6 & 49.4 & 40.0 \\ \hline
    f-VAEGAN-D2~\cite{f-vaegan-d2} & 70.6 & 57.6 & 63.5 & 60.1 & 48.4 & 53.6 & 38.0 & 45.1 & 41.3 \\ \hline
    LisGAN~\cite{LisGAN} & - & - & - & 57.9 & 46.5 & 51.6 & 37.8 & 42.9 & 40.2 \\ \hline
    CADA-VAE \cite{CADA-VAE} & 75.0 & 55.8 & 63.9 & 53.5 & 51.6 & 52.4 & 35.7 & 47.2 & 40.6 \\ \hline
    DASCN~\cite{DASCN} & - & - & - & 59.0 & 45.9 & 51.6 & 38.5 & 42.4 & 40.3 \\ \hline
    LsrGAN~\cite{LsrGAN} & - & - & - & 59.1 & 48.1 & 53.0 & 37.7 & 44.8 & 40.9 \\ \hline
    Zero-VAE-GAN~\cite{Zero-VAE-GAN} & 70.9 & 57.1 & 62.5 & 47.9 & 43.6 & 45.5 & - & - & - \\ \hline
    TF-VAEGAN~\cite{TF-VAEGAN} & 75.1 & 59.8 & 66.6 & 64.7 & 52.8 & 58.1 & 40.7 & 45.6 & {{43.0}} \\ \hline
    \hline
    DLFZRL~\cite{DLFZRL} & - & - & 60.9 & - & - & 51.9 & - & - & 42.5 \\ \hline
    RFF-GZSL~\cite{RFF-GZSL} & - & - & - & 56.6 & 52.6 & 54.6 & 38.6 & 45.7 & 41.9\\ \hline
    TCN~\cite{TCN} & 61.2 & 65.8 & 63.4 & 52.6 & 52.0 & 52.3 & - & - & - \\ \hline
    Disentangled-VAE~\cite{Disentangled-VAE} & 80.2 & 56.9 & 66.6 & 58.2 & 51.1 & 54.4 & 36.6 & 47.6 & 41.4 \\ \hline
    SE-GZSL~\cite{kim2021semantic} & 80.7 & 59.9 & {\underline{68.8}} & 60.3 & 53.1 & {{56.4}} & 40.7 & 45.8 & \underline{43.1} \\ \hline
    TDCSS~\cite{TDCSS} & 74.9 & 59.2 & 66.1 & 62.8 & 44.2 & 51.9 & - & - & - \\ \hline
    \hline
    DAZLE~\cite{DAZLE} & 75.7 & 60.3 & 67.1 & 59.6 & 56.7 & 58.1 & 24.3 & 52.3 & 33.2 \\ \hline
    Composer~\cite{Composer} & 77.3 & 62.1 & {\underline{68.8}} & 56.4 & 63.8 & {\underline{59.9}} & 22.0 & 55.1 & 31.4 \\ \hline
    
    \hline
    \rowcolor[HTML]{E0E0E0} 
    \textbf{ViT-GZSL (Ours)} & 84.8 & 63.3 & \textbf{72.5} & 78.2 & 59.9 & \textbf{67.9} & 46.1 & 51.7 & \textbf{48.8} \\
    \bottomrule
    \end{tabular}}
    \label{tab:result}
\end{table*}
\setlength{\tabcolsep}{2pt}

%% file: sections/05_conclusion.tex
In this paper, we proposed a new GZSL technique termed ViT-GZSL.
Key idea of ViT-GZSL is to employ ViT as an image feature extractor to maximize the attribute-related information included in the image feature.
In doing so, we can access the entire image region without the degradation of the image resolution, and together we can use the local image information preserved in the patch features.
To take full advantage of ViT, we used both the CLS feature and patch features in extracting the attribute-related image feature.
In particular, we proposed a novel attention-based module called AAM, which aggregates the attribute-related information contained in the patch features.
By exploiting both the AAM feature as well as CLS feature, the proposed ViT-GZSL achieved the state-of-the-art performance on AWA2, CUB, and SUN datasets.

%% file: sections/a0_appendix.tex
\section{Details of ViT}\label{appendix_sec:vit_details}

\begin{figure*}[t]
  \centering 
  \includegraphics[width=1.00\linewidth]{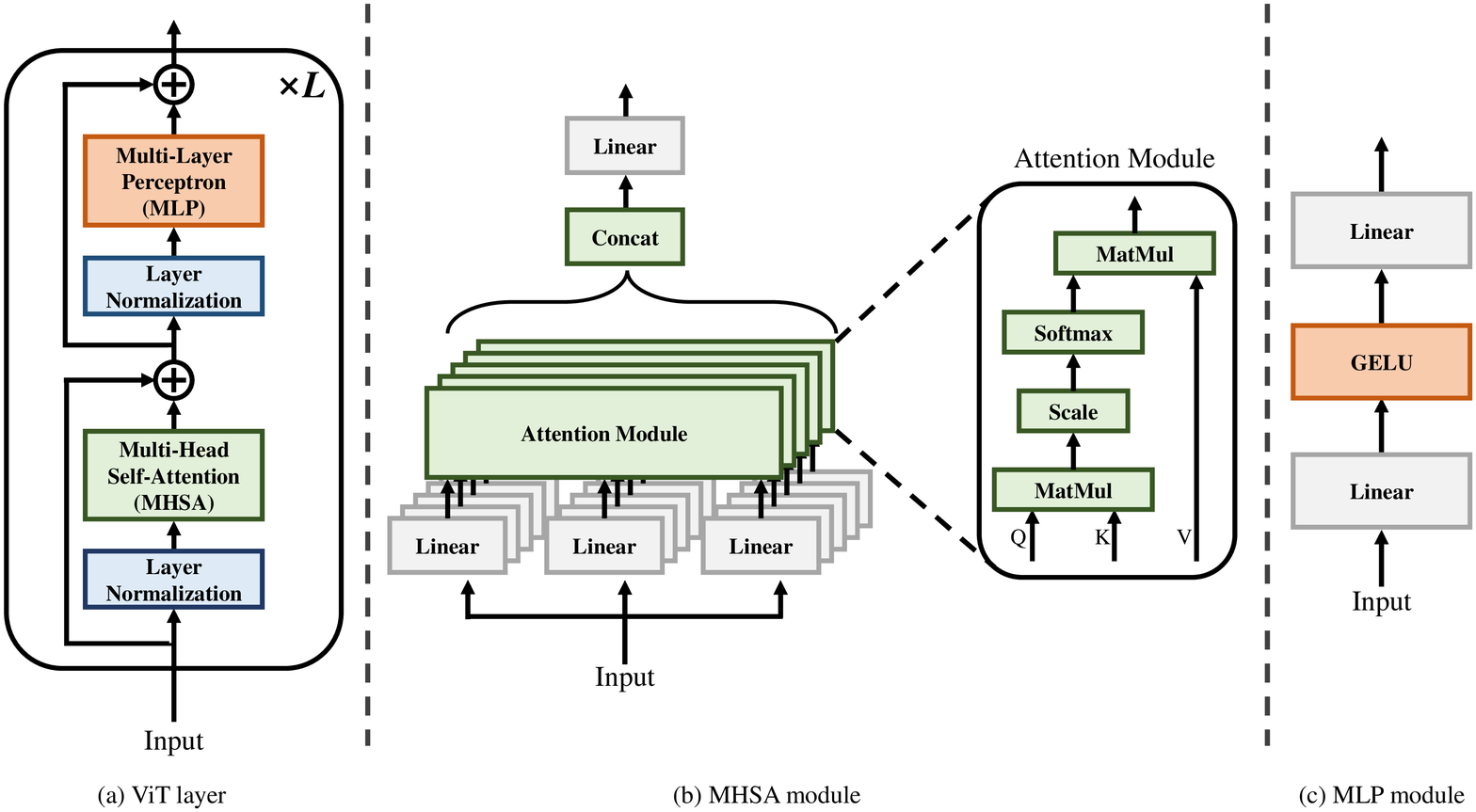}
  \caption{Illustration of the detailed architecture of ViT.}
  \label{fig:supply}
\vspace{-5pt}
\end{figure*}

ViT consists of multiple Transformer layers, each of which is composed of two modules, multi-head self-attention (MHSA) and multi-layer perceptron (MLP).
The input of each module first goes through the layer normalization (LN) and the output is added with the input through the residual connection (see Fig.~\ref{fig:supply}(a)).

As briefly explained in Section 3.1, the output $\mathbf{X}^{\ell}$ of the $\ell\text{-th}$ layer in ViT can be expressed as
\begin{align}
    \mathbf{X}^{\ell} &= \text{MLP}(\text{LN}(\mathbf{Z}^{\ell})) + \mathbf{Z}^{\ell}, \\
    \mathbf{Z}^{\ell} &= \text{MHSA}(\text{LN}(\mathbf{X}^{\ell-1})) + \mathbf{X}^{\ell-1}.
\end{align}
For simplicity, we define
\begin{align}
    \hat{\mathbf{X}}^{\ell-1} &=  \text{LN}(\mathbf{X}^{\ell-1}), \\
    \hat{\mathbf{Z}}^{\ell} &=  \text{LN}(\mathbf{Z}^{\ell}).
\end{align}

In the MHSA module, multiple self-attention operations, referred to as ``(attention) heads", are performed in parallel to capture the diverse relationship between input features. 
In the $h\text{-th}$ head, the input $\hat{\mathbf{X}}^{\ell-1}$ is first projected onto query $Q_h$, key $K_h$, and value $V_h$ using linear layers.
Then, the attention map $A_{h}$ is computed by a scaled dot-product of $Q_{h}$ and $K_{h}$ followed by the softmax operation:
\begin{align}
    Q_{h} &= \hat{\mathbf{X}}^{\ell-1} \mathbf{W}_{h,q} + \mathbf{b}_{h,q},   \\
    K_{h} &= \hat{\mathbf{X}}^{\ell-1} \mathbf{W}_{h,k} + \mathbf{b}_{h,k},  \\
    V_{h} &= \hat{\mathbf{X}}^{\ell-1} \mathbf{W}_{h,v} + \mathbf{b}_{h,v}, \\
    A_{h} &= \text{Softmax} \left ( Q_{h}K_{h}^{T} / \sqrt{D_{h}}\right ),
\end{align}
where $\mathbf{W}_{h,\{q,k,v\}}$ and $\mathbf{b}_{h,\{q,k,v\}}$ are weight and bias parameters for query, key, and value.
Here, $D_{h}=D/K$ is the dimension of each head where $D$ and $K$ are the dimension of input and the number of heads, respectively.
Then, the output $\text{SA}_{h}$ of the $h$-th head is expressed as a weighted sum of the rows in $V_{h}$:
\begin{align}
    \text{SA}_{h} &= A_{h} V_{h}.
\end{align}
Finally, the outputs of all attention heads are stacked together and then pass through another linear layer (see Fig.~\ref{fig:supply}(b)):
\begin{align} 
    \text{MHSA}(\hat{\mathbf{X}}^{\ell-1}) &= [\text{SA}_{1};\cdots ;\text{SA}_{K}] \mathbf{W}_{mhsa} + \mathbf{b}_{mhsa}.
\end{align}

The MLP module consists of two linear layers with the GELU intermediate activation function (see Fig.~\ref{fig:supply}(c)):
\begin{align} 
    \text{MLP}(\hat{\mathbf{Z}}^{\ell}) \hspace{-.1cm}&=\hspace{-.1cm} \left( \mathbf{\phi} \left( \hat{\mathbf{Z}}^{\ell} \mathbf{W}_{mlp, 1} + \mathbf{b}_{mlp, 1} \right) \right) 
    \mathbf{W}_{mlp, 2}  + \mathbf{b}_{mlp, 2} ,
\end{align} 
where $\mathbf{\phi}$ is GELU activation function, 
$\mathbf{W}_{mlp,\{1,2\}}$ are weight parameters, 
and $\mathbf{b}_{mlp\{1,2\}}$ are bias parameters.

\section{ViT Layer Selection on Other Dataset}\label{appendix_sec:layer_selection}

\begin{figure}[t]
    \centering 
    \includegraphics[width=0.99\linewidth]{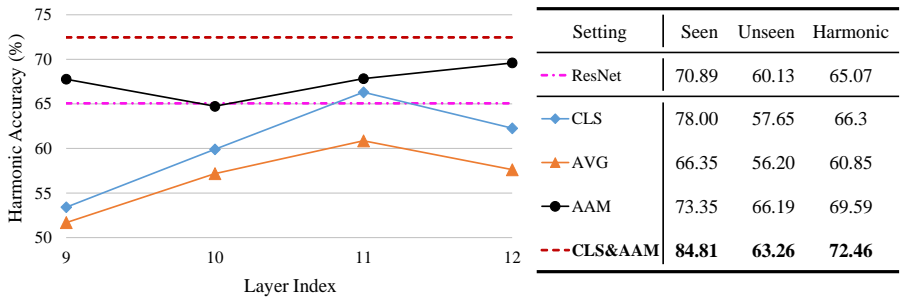}
    \vspace{-5pt}
    \caption{GZSL performance evaluated on the AWA2 dataset.
    }
    \label{fig:layer_accuracy_table_awa2}
\end{figure} 

\begin{figure}[t]
    \centering 
    \includegraphics[width=0.99\linewidth]{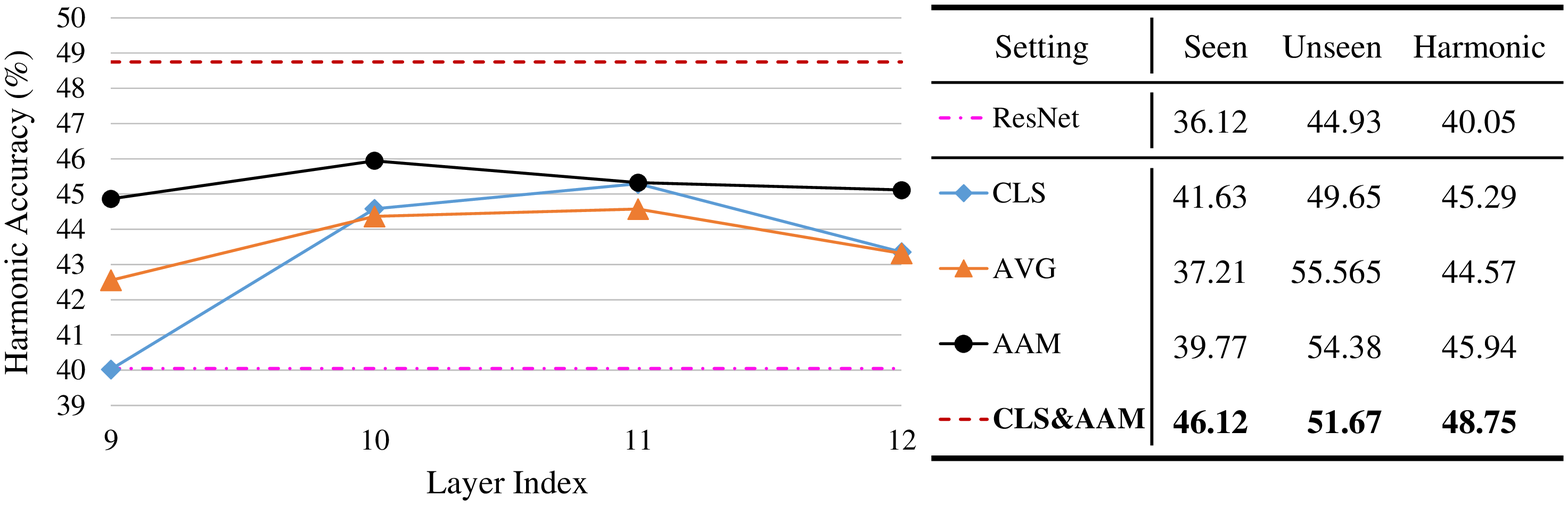}
    \vspace{-5pt}
    \caption{GZSL performance evaluated on the SUN dataset.
    }
    \label{fig:layer_accuracy_table_sun}
\end{figure}

As in Section 4.2, we demonstrate the effect on the accuracy with respect to the Transformer layer used in the feature extraction.
In Fig.~\ref{fig:layer_accuracy_table_awa2} and~\ref{fig:layer_accuracy_table_sun}, we summarize the results on the AWA2 and SUN datasets, respectively.
For AWA2, we can observe that the AVG feature performs much worse than both CLS and AAM features, and the accuracy gap between CLS and AAM is noticeable.
On the other hand, the gap between CLS and AAM in SUN dataset is not as large as in the AWA2 dataset.
We assume that this is because most of the attributes of the SUN dataset are related to the global information (e.g. `cluttered space’ and `open area' attributes).
We would like to emphasize that the proposed ViT feature (concatenation of CLS and AAM feature) always achieves the best performance since the attribute-related information, not captured by the CLS feature, can be compensated by using the AAM feature together. 

\section{Additional Training Details}\label{appendix_sec:training_details}

\subsection{Softmax Classifier for GZSL Classification}
\label{subsec:softmax classifier}

As mentioned in Section 3.3, we use the simple softmax classifier as the image feature classifier to directly compare the ViT feature with the conventional ResNet feature.
In the softmax classifier, the probability that an image feature $\mathbf{x}$ belongs to class $y$ is expressed as
\begin{equation}
    \mathcal{P}(y|\mathbf{x})
    = \frac{\exp{(\mathbf{w}_{y}^T \mathbf{x} + b_{y})}}{\sum_{c \in Y_{s} \cup Y_{u}} \exp{(\mathbf{w}_{c}^{T} \mathbf{x} + b_{c})}},
\end{equation}
where $Y_{s}$ is the set of seen classes, $Y_{u}$ is the set of unseen classes, and $\mathbf{w}_{c}$ and $b_{c}$ are parameters to be learned.
Let $\mathbf{x}_{s}$ be the image feature of seen classes $s \in Y_{s}$ and $\widetilde{\mathbf{x}}_{u}$ be the synthetic image feature of unseen classes $u \in Y_{u}$, then the softmax classifier is trained to minimize the cross-entropy loss:
\begin{equation}
    \mathcal{L}_{\text{cls}} = - \hspace{-.2cm} \sum_{s \in Y_{s}} \hspace{-.08cm} \sum_{i=1}^{N_{s}} \log \mathcal{P}(s|\mathbf{x}^{i}_{s}) - \hspace{-.2cm} \sum_{u \in Y_{u}} \hspace{-.08cm} \sum_{i=1}^{N_{u}} \log \mathcal{P}(u|\widetilde{\mathbf{x}}_{u}^{i}), 
\end{equation}
where $N_{s}$ is the number of image features of seen classes $s \in Y_{s}$ and $N_{u}$ is the number of generated image features of unseen classes $u \in Y_{u}$.

\subsection{Auxiliary Classifier for the AAM and F2A Training}
\label{subsec:AAM classifier}

As mentioned in Section 3.2, we use the auxiliary classifier to train AAM.
The classifier is jointly trained with AAM such that the correct class label can be predicted from the AAM feature.
The classifier consists of leaky-ReLU and linear layer, followed by the softmax layer (see the dashed box in Fig. 4).
In the training phase, we utilize the cross-entropy (CE) loss between the predicted probabilities and the seen class labels.

Note that the AAM and F2A are jointly trained.
The overall loss function for AAM feature extraction is expressed as 
\begin{equation}
    \mathcal{L}_{\text{AAM,F2A}}= \sum_{s \in Y_{s}} \sum_{i=1}^{N_{s}} \| a_s - \hat{a}^i_s \|^2 -\sum_{s \in Y_{s}} \sum_{i=1}^{N_{s}} \log \mathcal{P} \big( s | \mathbf{x}_{\text{AAM}, s}^i \big),
\end{equation}
where the first term is MSE loss for F2A module, and the second term is CE loss for both AAM and F2A module.
$\hat{a}^i_s$ indicates the synthetic attribute which is generated from F2A module using the CLS feature $\mathbf{x}^i_{\text{CLS},s}$ as input.

\subsection{CVAE for Unseen Feature Generation}\label{appendix_sec:elbo}

In synthesizing ViT features from image attributes, CVAE employs two networks, an encoder and a decoder, which are trained to maximize the conditional likelihood $p(\mathbf{x}_{\text{ViT}} | \mathbf{a})$. 
In the training phase, the encoder maps the ViT feature $\mathbf{x}_{\text{ViT}}$ and the image attribute $\mathbf{a}$ into a random Gaussian latent vector $\mathbf{z} \sim \mathcal{N}(\mathbf{\mu}_{\mathbf{z}}, \mathbf{\Sigma}_{\mathbf{z}})$.
Then, the decoder reconstructs the original ViT feature $\mathbf{x}_{\text{ViT}}$ using the latent vector $\mathbf{z}$ and the image attribute $\mathbf{a}$. 
Let $\widehat{\mathbf{x}}_{\text{ViT}}$ be the reconstructed ViT feature, then the encoder and decoder are jointly trained to minimize the evidence lower bound (ELBO) of the conditional log likelihood $\log p(\mathbf{x}_{\text{ViT}} | \mathbf{a})$~\cite{VAE}:
\begin{align} 
    \mathcal{L}_{\text{CVAE}}(\mathbf{x}_{\text{ViT}}, \mathbf{a})
    =& \mathcal{L}_{\text{recon}}(\mathbf{x}_{\text{ViT}},\widehat{\mathbf{x}}_{\text{ViT}}) +KL\big(\mathcal{N}(\mathbf{\mu}_{\mathbf{z}}, \mathbf{\Sigma}_{\mathbf{z}}),\mathcal{N}(\mathbf{0}, \mathbf{I}) \big),
    \label{eq:CVAE loss}
\end{align}
where $\mathcal{L}_{recon}(\mathbf{x}_{\text{ViT}},\widehat{\mathbf{x}}_{\text{ViT}})$ is the reconstruction error and $KL\big(\mathcal{N}(\mathbf{\mu}_{\mathbf{z}}, \mathbf{\Sigma}_{\mathbf{z}}),\mathcal{N}(\mathbf{0}, \mathbf{I}) \big)$ is the Kullback-Leibler (KL) divergence between the latent vector distribution and the standard Gaussian distribution.

After the CVAE training, we use the decoder in synthesizing ViT features of unseen classes.
Specifically, for each unseen class $u$, we generate synthetic ViT features $\widetilde{\mathbf{x}}_{\text{ViT}}$ by sampling a latent vector $\mathbf{z} \sim \mathcal{N}(\mathbf{0}, \mathbf{I})$ and then exploiting $\mathbf{z}$ and the image attribute $\mathbf{a}_{u}$ as inputs to the decoder.
By resampling the latent vector, a sufficient number of synthetic ViT features can be generated.